\documentclass[runningheads]{llncs}
\usepackage{cite}
\usepackage{graphicx}
\usepackage{amsmath}

\usepackage{verbatim}
\usepackage{multirow}
\usepackage{amssymb}
\usepackage{caption}
\usepackage{floatrow}
\usepackage{subfigure}
\usepackage{multirow}
\usepackage{comment}
\usepackage{siunitx}
\usepackage{booktabs}
\usepackage{color}
\usepackage{hyperref}
\hypersetup{
    colorlinks=true,
    linkcolor=blue,
    filecolor=magenta,      
    urlcolor=red,
}

\usepackage{bm}

\newfloatcommand{capbtabbox}{table}[][\FBwidth]
\begin{document}
\title{Sequential Learning on Liver Tumor Boundary Semantics and Prognostic Biomarker Mining}
\author{Jieneng Chen\textsuperscript{1}, Ke Yan\textsuperscript{2}, Yu-Dong Zhang\textsuperscript{3},
Youbao Tang\textsuperscript{2}, \\Xun Xu\textsuperscript{3}, Shuwen Sun\textsuperscript{3}, Qiuping Liu\textsuperscript{3}, Lingyun Huang\textsuperscript{4}, Jing Xiao\textsuperscript{4}, \\Alan L. Yuille\textsuperscript{1}, Ya Zhang\textsuperscript{5}, and Le Lu\textsuperscript{2}}

\authorrunning{J. Chen et al.}

\institute{
\textsuperscript{1}Johns Hopkins University\\
\textsuperscript{2}PAII Inc.\\
\textsuperscript{3}The First Affiliated Hospital of Nanjing Medical University\\
\textsuperscript{4}Ping An Technology \\
\textsuperscript{5}Shanghai Jiao Tong University\\ }

\titlerunning{Tumor Boundary Semantics and Prognostic Biomarker Mining}
%
%
\maketitle              
\begin{abstract}

The boundary of tumors (hepatocellular carcinoma, or HCC) contains rich semantics: capsular invasion, visibility, smoothness, folding and protuberance, etc. Capsular invasion on tumor boundary has proven to be clinically correlated with the prognostic indicator, microvascular invasion (MVI). Investigating tumor boundary semantics has tremendous clinical values. In this paper, we propose the first and novel computational framework that disentangles the task into two components: spatial vertex localization and sequential semantic classification. 
(1) A HCC tumor segmentor is built for tumor mask boundary extraction, followed by polar transform representing the boundary with radius and angle. Vertex generator is used to produce fixed-length boundary vertices where vertex features are sampled on the corresponding spatial locations. 
(2) The sampled deep vertex features with positional embedding are mapped into a sequential space and decoded by a multilayer perceptron (MLP) for semantic classification. Extensive experiments on tumor capsule semantics demonstrate the effectiveness of our framework. Mining the correlation between the boundary semantics and MVI status proves the feasibility to integrate this boundary semantics as a valid HCC prognostic biomarker.

\end{abstract}
\section{Introduction}
Microvascular invasion (MVI) has been clinically identified as a prognostic factor of hepatocellular carcinoma (HCC) after surgical treatment, whereas it is undetectable preoperatively on diagnostic imaging \cite{5shah2007recurrence,6chan2018development}. Microscopic features of HCC such as tumor size, capsule and margin are hypothesized as important predictors of MVI \cite{8an2019imaging,9zhu2019incomplete}. Tumor capsule, specific for hepatocarcinogenesis, was observed in 70\% of progressed HCC \cite{10kojiro2005histopathology}. Histologically, tumor capsule contains two layers: the inner layer is composed of tight fibrous tissue containing thin, slit-like vascular channels, and the outer layer is composed of looser fibrovascular tissue \cite{11choi2014ct}. 
Tumor capsule invasion increases the risk of vascular invasion and intrahepatic metastasis, generally indicating poor cancer patient prognosis. 
However, despite its strong potential, assessing the presence and integrity of radiological HCC capsule is an expert-based subjective evaluation. Inter- and intra-observer variations and lack of reproducibility are the major roadblocks limiting its wide adoption \cite{14ehman2016rate}. There is a critical unmet need to develop new yet effective computational methods and means to objectively and quantitatively  examine the capsular invasion and focal extensional nodule to predict the precision prognosis of patients with HCC. 

\begin{figure*}[t!]
    \centering
    \includegraphics[width=0.98 \textwidth]{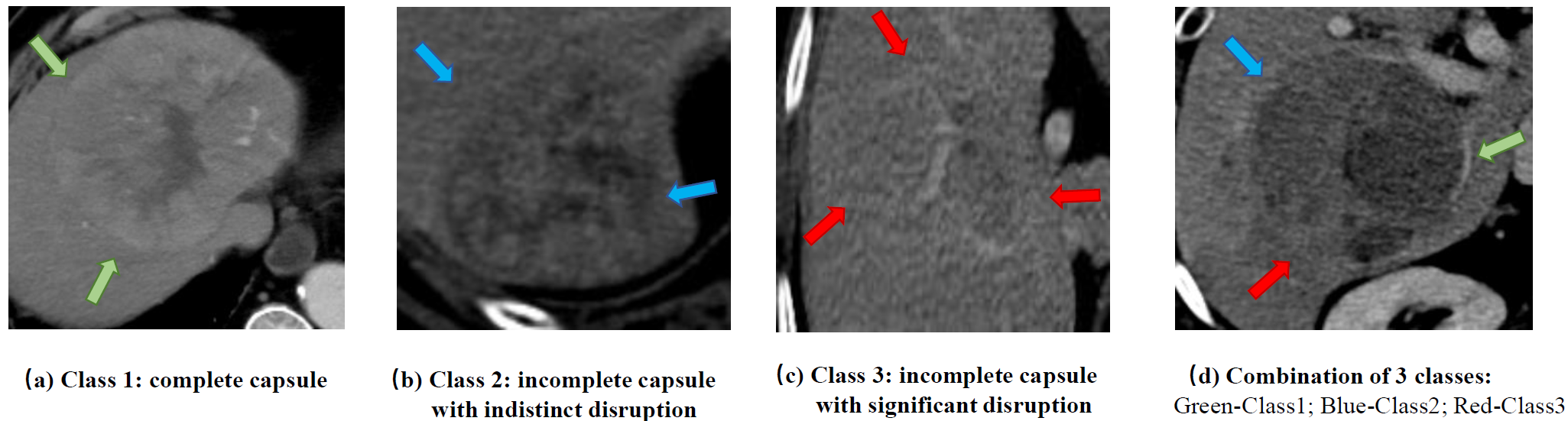}
    \caption{Categories of tumor capsule invasion as a type of boundary semantics. 
    }
    \label{fig:intro}
\end{figure*}

The capsule in radiology can be interpreted as the semantics on tumor boundary (See Fig. \ref{fig:intro}), and this problem falls into learning the tumor boundary semantics, i.e., dense classification on the boundary pixels. 
We propose a novel framework that disentangles the task into two pillars: the spatial localization and the sequential classification. (1) The spatial localization aims to precisely identify the vertices on tumor boundary that is extracted from predicted tumor mask. A polar coordinate transform is used to represent the boundary with radius and angle so that the boundary can be divided to $N$ grids with equidistant angle. Then we perform an efficient vertex generator to produce the localized vertices' coordinates. (2) Our sequential learning tackles  the semantic classification on the localized vertices' coordinates via sampled deep vertex features (that are concatenated with positional embedding). Finally the formed sequence features are decoded by a multilayer perceptron (MLP) for semantic classification.

To the best of our knowledge, this is the first work to solve the dense tumor boundary semantics mining problem and formulate it in a sequential learning manner. The polar coordinate transform enables us to obtain spatially uniform boundary coordinates. The vertex sequential features are sampled from multi-scale pyramid features, permitting to naturally integrate low-/mid-/high-level cues on boundary semantics. We demonstrate the effectiveness of our approach on
two tumor boundary semantic datasets: capsular invasion (CAP) and focal extensional nodule (FEN). Our method improves the baseline of entangled optimization by 23.26\% F1 score on CAP dataset and 10.32\% F1 score on FEN dataset. Moreover, we conduct a study of prognostic biomarker mining to validate the clinical correlation between boundary semantics and MVI status.

\noindent\textbf{Previous work.}
UPI-Net is proposed in \cite{qi2019upi} to detect semantic contour in Placental Ultrasound by binary boundary segmentation which is also studied in nature images by \cite{hariharan2011semantic,shen2015deepcontour,bertasius2015high}. In contrast, we detect multi-classes boundary semantics independent to object semantic and/or instance category. Polar representation is used \cite{schmidt2018cell} to localize cell via star-convex polygons and to model foreground instances for instance segmentation~\cite{xie2020polarmask,xu2019explicit}. Sequential learning is a common task in natural language processing such as machine translation \cite{sutskever2014sequence, vaswani2017attention}, and various vision tasks \cite{mao2014deep,chen2019flgr,dosovitskiy2020image,chen2021transunet}. Moreover, to tackle with the class imbalance problem in sequential learning, Li \emph{et al.}~\cite{li2019dice} introduced a sequential dice loss into NLP tasks as the training objective, which is adopted in our work. 

\section{Method}
\label{sec:method}
\begin{figure*}[t!]
    \centering
    \includegraphics[width=0.85\textwidth]{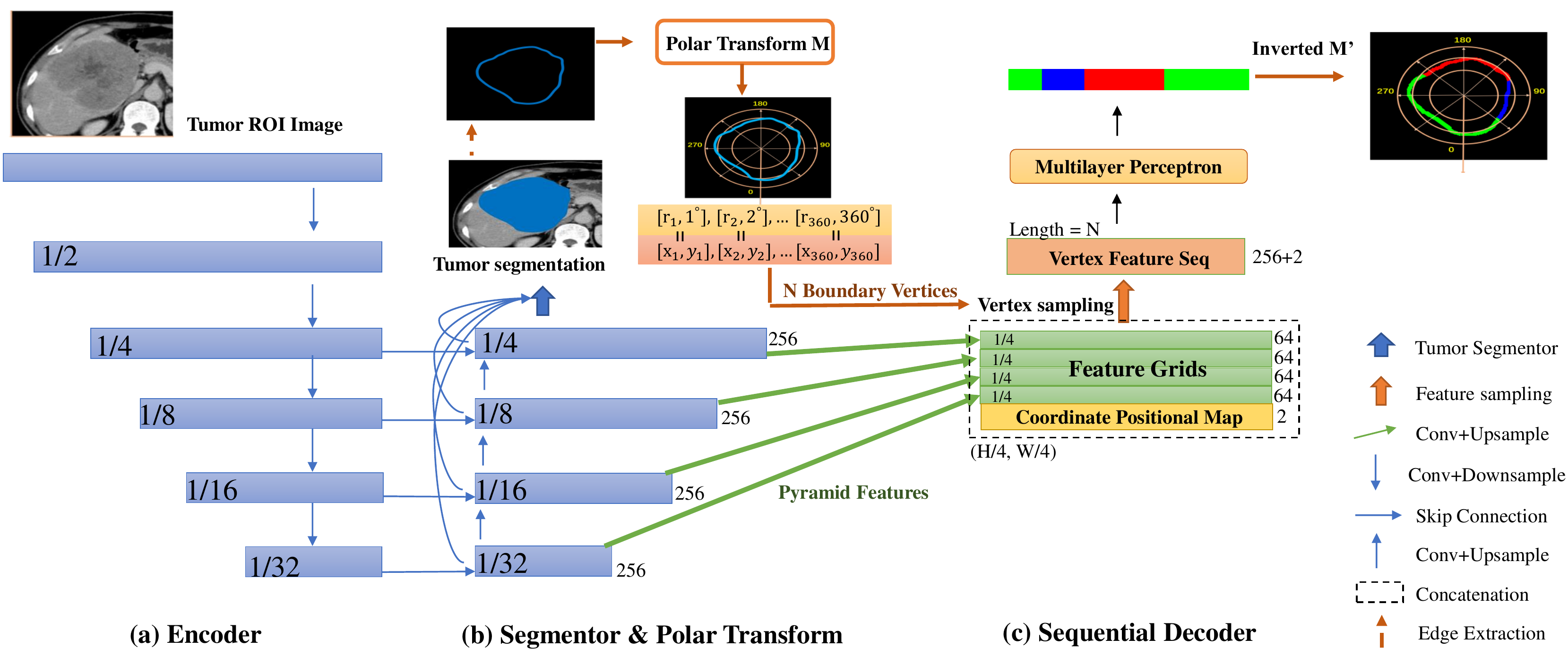}
    \caption{Overview of our tumor boundary semantics modeling framework.
    }
    \label{fig:framework}
\end{figure*}

Given a tumor RoI image $\bm{\mathrm{x}} \in \mathbb{R}^{H \times W \times C}$, our goal is to predict the corresponding pixel-wise label map along tumor boundary. Unlike existing approaches of directly training a segmentation network (\emph{e.g.}, U-Net), our method converts the problem into conducting the sequential prediction on a 1D band label map $\bm{\mathrm{y}} \in \mathbb{R}^{N_{angle}}$.
Our overall framework is depicted in Fig. \ref{fig:framework}.  

\vspace{-2mm}

\subsection{Tumor Boundary Spatial Localization}
~\label{sec:spatial}

Our encoder is initialized with a ResNet-50 network~\cite{he2016deep}. ResNet features in scale $\bm{\mathrm{S}} \in \{\frac{1}{4}, \frac{1}{8}, \frac{1}{16}, \frac{1}{32}\}$ will pass through the U-shape blocks, where each block consists of a 2×upsampling operator, a skip-connection, a 3×3 convolution layer, and a ReLU layer successively. Multi-scale pyramid features are generated with each scale feature $\bm{\mathrm{x}} \in \mathbb{R}^{\frac{H}{S} \times \frac{W}{S} \times 256}$. We first build a tumor segmentation model (See ) on top of the multi-scale pyramid features that upsample every scale feature into scale 1/4 and merge them with an add operator, followed by a 3×3 convolution layer, 4× upsampling operator and a softmax activation layer to predict the binary tumor mask $\bm{\mathrm{y_t}} \in \mathbb{R}^{H \times W}$. This tumor segmentor is trained with Dice loss and cross-entropy loss. After obtaining the tumor mask, we employ the residual of the dilation and the erosion of predicted tumor mask as in \cite{tang2020net} to generate the tumor boundary. Gaussian blurring with a 5$\times$5 kernel is used to make the boundary thickness closer to the human annotation.

Polar coordinates is a two-dimensional coordinate system where each point on a plane is determined by a distance from a reference point and an angle from a reference direction. The reference point (analogous to the origin of a Cartesian coordinate system) is called the pole, and the ray from the pole in the reference direction is the polar axis. The distance from the pole is called the radial coordinate, and the angle is called the angular coordinate. Hence, $(x_{i}, y_{i})$ in Cartesian coordinate system is denoted as $(r_{i}, \theta_{i})$ in polar coordinate system $M$; the pole (0, 0) in polar system is exactly the centroid of tumor mask $(x_{c}, y_{c})$ in Cartesian system. 
Cartesian position of the vertex can be recovered from the inverted transform $M'$, where $x_{i} =  r_i \cos\theta_{i} + x_{c}$ and $y_{i} =  r_i \sin\theta_{i} + y_{c}$.

We propose an efficient vertex generator to produce $N$ boundary vertices $(x_1, y_1), \ldots, (x_N, y_N)$. $N$ rectangle grids in polar coordinate system standing for $N$ rays  with equidistant angle $\Delta\theta=\frac{360^{\circ}}{N}$ are generated. Specifically, grid $k \in \{1, 2, \ldots, N\}$ is filled with a set of candidate vertices $(r_{g}, \theta_{g}) \in \mathbb{G}^{R \times 3}$ in polar representation, where $\theta_{g} \in \{(k-1)*\Delta\theta, k*\Delta\theta, (k+1)*\Delta\theta\}$, $r_{g} \in \{r|\theta=\theta_g\}$, and $R$ approximates to the $\mathop{\max} {(r|\theta=k*\Delta\theta)} - \mathop{\min} {(r|\theta=k*\Delta\theta)}$. 
For robustness, we randomly sample a vertex point each time in $N$ grids iteratively as a data augmentation strategy.

\subsection{Deep Sequential Learning on Tumor Semantics}
~\label{sec:seq}

There are four scales of features from the feature pyramid (in Sec. \ref{sec:spatial}) being processed to the same feature size $\frac{H}{4} \times \frac{W}{4} \times 64$ with bilinear upsampling and 3$\times$3 convolution, and concatenated channel-wise to generate the feature $\bm{\mathrm{x_p}} \in \mathbb{R}^{\frac{H}{4} \times \frac{W}{4} \times 256}$. To retain the positional information of each tumor boundary vertex and to ensure the structural relationship to be learned, we make use of the coordinate positional map $\bm{\mathrm{x_{coor}}} \in \mathbb{R}^{\frac{H}{4} \times \frac{W}{4} \times 2}$, where the channels representing the $x$ and $y$ Cartesian coordinates are normalized to $[-1, 1]$. Last we concatenate $\bm{\mathrm{x_p}} $ and $\bm{\mathrm{x_{coor}}} $ channel-wise to form the feature grids $\bm{\mathrm{x_g}} \in \mathbb{R}^{\frac{H}{4} \times \frac{W}{4} \times 256+2}$. Given an arbitrary point $(x_i, y_i)$, a corresponding grid of feature $ \in \mathbb{R}^{258}$ can be sampled from $\bm{\mathrm{x_g}}$. Recall that each time the efficient vertex generator will generate a set of boundary vertices $(x_1, y_1), (x_2, y_2), \ldots, (x_N, y_N)$, where total $N$ grids of feature will be sampled from $\bm{\mathrm{x_g}}$ and further sequentialized to $\bm{\mathrm{x_{seq}}} \in \mathbb{R}^{N \times 258}$. A multilayer perceptron (MLP) is adopted to decode $\bm{\mathrm{x_{seq}}} \in \mathbb{R}^{N \times 258}$ into 1D sequential classification prediction of $f\left(\mathbf{x}\right) \in \mathbb{R}^{N}$. The MLP contains two layers with a GELU non-linearity and a softmax activation. 
To alleviate the class imbalance where statistically the first class makes accounts for 70\%, we exploit the 1D sequential dice loss~\cite{li2019dice} with cross entropy loss as the training objective for the sequential decoding. Let $\mathcal{D}$ be the dataset and the labeled data pair $\mathbf{x}_{i}, \mathbf{y}_{i} \in \mathcal{D}$, the sequential loss $\mathcal{L}_{seq}$ is formulated as:
\begin{equation}\label{equ:loss_rec}
    \mathcal{L}_{seq} = \underbrace{1-\sum_{\mathbf{x}_{i}, \mathbf{y}_{i} \in \mathcal{D}} 2 \frac{\sum f\left(\mathbf{x}_{i}\right)\mathbf{y}_{i}}{\sum f\left(\mathbf{x}_{i}\right) + \sum \mathbf{y}_{i}}}_{1D~dice~loss}~\underbrace{-\sum_{\mathbf{x}_{i}, \mathbf{y}_{i} \in \mathcal{D}}\mathbf{y}_{i}\cdot\log\left( f\left(\mathbf{x}_{i}\right) \right)}_{cross~entropy~loss}
\end{equation}

\subsection{Prognostic Tumor Biomarker Mining} \label{sec:prog}
Microvascular invasion (MVI) score is a significant prognostic indicator of HCC in pathological imaging findings. We aim to fully exploit the correlation between MVI and our radiological imaging measurements. 
For each patient with multiple slices of images, we inference the data using the algorithm described above in a slice-by-slice fashion and stack the results as the prediction. The number of pixels of each class in the 3D tumor boundary is counted and divided by the total number of pixels to obtain a feature vector with three variables (e.g. [0.6, 0.1, 0.3]), where two of them are independent variables making up the patient-specific capsular biomarker. A logistic regression (LR) classifier is employed to analyze the correlation between the capsular biomarker and MVI.

\section{Experiments and Discussion}

\noindent\textbf{Dataset collection.}
A total of 358 unique patients ($4049$ axial slices) with pathologically confirmed liver tumor (HCC) were included in our study, which are split to 193: 61: 104 for training, validation and testing, respectively. All patients underwent standard multi-phase contrast-enhanced abdominal CT imaging within 2 weeks before surgery. 
\textit{Capsular invasion} \textbf{(CAP)} was categorized using the following three-point scale: (1) complete capsule or invisible capsule with smooth tumor margin; (2) incomplete capsule with indistinct disruption; or (3) incomplete capsule with significant disruption. We also collect a type of tumor boundary semantics named \textit{focal extensive nodule} \textbf{(FEN)} to measure the degree of protruding into the non-tumor parenchyma. FEN was assessed via a three-point criterion: (1) smooth margin without FEN; (2) slight FEN (the number of FEN is less than 3); or (3) significant FEN (the number of FEN is 3 or more). CAP and FEN classes were labeled manually using the referring three-point criteria by three board-certified radiologists on 5mm portal-venous CT images slice-by-slice. For the prognostic biomarker mining, we collected the patient-level labels of \textbf{microvascular invasion (MVI)} from their associated histopathologic examinations, where 0 stands for MVI negative; 1 for MVI positive. For inter-reader variability analysis, a subset from test-set consisting of 63 patients with 591 axial slices are labelled repeatedly by 3 radiologists Y2, Y4 and Y10, whose year-of-practice are 2, 4 and 10 respectively.

\noindent\textbf{Evaluation metrics.}
 To evaluate the sequential classification, we perform five quantitative measures including F1 score, accuracy, AUC score, precision and recall. We adopted Dice Similarity coefficient (DSC) to evaluate the tumor segmentation accuracy that is associated with boundary extraction quality. Following \cite{yao2020deepprognosis}, we perform quantitative measures including sensitivity, specificity, AUC, and Youden's J index (J) to evaluate the logistic regression model of MVI prediction. The class weights in all LR models are adjusted to maximize Youden's J index of (sensitivity+specificity-1).

\noindent\textbf{Implementation details.} For all experiments, we apply simple data augmentations, e.g., random rotation, random resize and flipping. Models are trained with SGD optimizer with learning rate 0.01, momentum 0.9 and weight decay 1e-4. The default batch size is 64 and default training epoch is 300. The tumor segmentation branch and the sequential classification branch are optimized jointly in training. In addition, during training, we use the proposed train-time efficient vertex generator while during testing we directly generate the vertices from predicted tumor boundary.

\begin{figure*}[t!]
    \centering
    \includegraphics[width=0.92\textwidth]{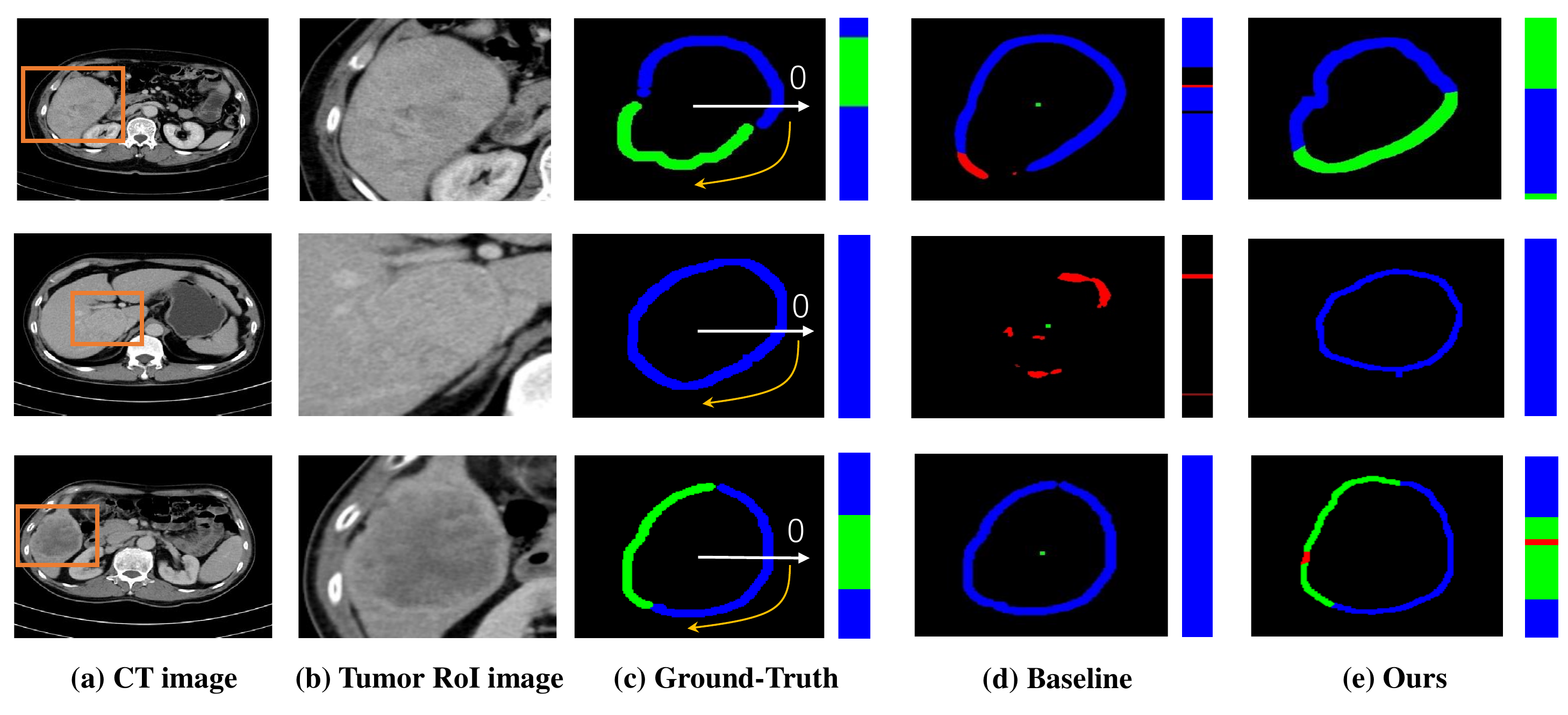}
    \caption{Qualitative comparison. (a) Whole CT image, (b) Tumor RoI image, (c) Ground-Truth, (d) Baseline prediction, (e) Our prediction. The sequence is flatten clockwise with the starting ray of 0 angle shown in (c).
    }
    \label{fig:vis} \vspace{-4mm}
\end{figure*}

\noindent\textbf{Quantitative comparisons.}
The baseline mentioned in Sec. \ref{sec:method} is a multi-classes segmentation network consisting of the same ResNet-50 encoder as ours and an UNet segmentation decoder. The baseline directly produces 2D semantic boundary mask which is further converted to 1D sequential band prediction using the vertex-based mask sampling, similar to our test-time vertex generating and feature sampling. There are two branches in our framework: tumor segmentation and sequential prediction. Table. \ref{tab:main} shows our framework can achieve 87.74\% tumor DSC on the capsular invasion (CAP) dataset and 88.29\% tumor DSC on the focal extensional nodule (FEN) dataset, facilitating high-quality boundary extraction and vertices generation. For sequential prediction, our framework outperforms the baseline on all measures, where the F1 score is superior by 23.25\% on CAP and by 10.24\% on FEN dataset. The average tumor segmentation accuracy achieves 88\% DSC, indicating there is a margin of error between predicted and ground-truth tumor boundary. An upper-bound performance of our overall framework may be expected by replacing the predicted boundary with ground-truth in vertex localization. However, the quantitative results  in Table. \ref{tab:upper} ($\textit{GT\_Vertices}$) indicate otherwise. This observation can be explained from \cite{hou2013boundary} that claims the reliability of human-labelled boundary ground-truth is questionable due to ill-posed nature of boundary detection and uncertainty caused by human-annotated error. This result suggests that our framework is robust against slight shifts of boundaries due to its built-in efficient vertex generator and pyramid features. 
We also provide {\bf qualitative comparison} results on the Capsular invasion prediction, as shown in Fig~. \ref{fig:vis}. From there, our model is observed as the better solution to recover the boundary shape while the baseline fails sometimes (in the second row); and has stronger representation power to encode the context and distinguish the boundary semantics.

\begin{table}[t] 
\centering
\footnotesize
\renewcommand\arraystretch{1.1}
\begin{tabular}{cc|ccccc|c}
\toprule
Dataset              & Model    & \textit{F1} & \textit{Accuracy}   & \textit{AUC} & \textit{Precision} & \textit{Recall} & \textit{DSC-Tumor} \\ \hline \hline
\multirow{2}{*}{CAP} & Baseline & 25.28                   & 47.57 & 51.48                    & 33.09     & 23.99 & -  \\ \cline{2-8} 
                     & Ours     & \textbf{48.42}          & \textbf{64.19} & \textbf{64.85}   & \textbf{54.15}     & \textbf{48.23} & 87.74  \\ \hline 
\multirow{2}{*}{FEN} & Baseline & 26.26                   & 44.68 & 49.36                    & 34.58     & 25.62 & - \\ \cline{2-8} 
                     & Ours     & \textbf{36.58}          & \textbf{52.37} & \textbf{59.59}                    & \textbf{42.52}     & \textbf{37.46} & 88.29 \\ 
\toprule \vspace{-5mm}
\end{tabular}

\caption{Comparisons on CAP dataset and FEN dataset. The evaluation on sequential classification are based on F1, accuracy, AUC, precision and recall(\%). We use DSC(\%) to evaluate the tumor segmentation in our method.}
\label{tab:main}
\end{table}

\noindent\textbf{Comparison with inter-reader variability.} To quantify the inter-reader variability issue and how our approach measures against it, we compare the three radiologists' annotations (Y2, Y4, Y10) with each other and our prediction against them. The relatively poor inter-reader consistency in Table~\ref{tab:inter} show that the task is intrinsically challenging for human readers and the boundary semantics annotations lack of objectivity. Taking the ten-year practicing radiologist (Y10) as standard, our automated performance is closely equivalent to the four-year experienced radiologist (Y4), clearly proving the usability of our algorithm. 

\begin{table}[t] 
\centering
\footnotesize
\renewcommand\arraystretch{1.0}
\begin{tabular}{cc|ccccc}
\toprule
R1           & R2          & \textit{F1} & \textit{Accuracy} & \textit{AUC} & \textit{Precision} & \textit{Recall} \\ \hline \hline
Y2            & Y10         & 51.40       & 59.18             & 66.05        & 55.98              & 51.78           \\
Y2            & Y4          & 54.86       & 62.23             & 65.78        & 57.04              & 55.80           \\
Y4            & Y10         & 49.87       & 57.92             & 63.74        & 50.87              & 53.86           \\
\multicolumn{2}{c}{average} & 52.04       & 59.78             & 65.19        & 54.63              & 53.81           \\ \hline
Ours          & Y2          & 38.78       & 54.16             & 62.52        & 38.00              & 46.28           \\
Ours          & Y4          & 42.49       & 56.67             & 63.15        & 41.70              & 50.20           \\
\textbf{Ours}          & \textbf{Y10}         & 48.60       & 67.65             & 65.24        & 48.72              & 53.26           \\
\multicolumn{2}{c}{average} & 43.29       & 59.49             & 63.64        & 42.81              & 49.91      \\\toprule \vspace{-5mm}
\end{tabular}
\caption{Inter-reader variability analysis. R1 and R2 are annotation providers. Taking the ten-year practiced radiologist (Y10) as the standard, our algorithm's performance is comparable to the four-year practiced radiologist's (Y4).}
\label{tab:inter} \vspace{-4mm}
\end{table}

\begin{table}[t] 
\centering
\footnotesize
\renewcommand\arraystretch{1.1}
\begin{tabular}{c|c|c|c|ccccc}
\toprule
\textit{PyraFeat} & \textit{CoordPos} & \textit{N\_Vertices} & \textit{GT\_Vertices} & \textit{F1}    & \textit{Accuracy}   & \textit{AUC}   & \textit{Precision} & \textit{Recall} \\
\hline\hline
\checkmark    &             & 90        &              & 44.99          & 63.28          & 64.2           & 51.15              & 44.47           \\
               & \checkmark & 90        &              & 41.44          & 61.76          & 63.69          & 48.12              & 40.52           \\
\checkmark     & \checkmark & 30        &              & 47.84          &  62.71         & 63.94          & 54.19              & 47.44         \\
\checkmark     & \checkmark & 90        & \checkmark   & 47.5           & \textbf{64.23} & 64.79          & 53.63              & 47.23           \\
\checkmark     & \checkmark & 90        &              & \textbf{48.42} & 64.19          & \textbf{64.85} & \textbf{54.15}     & \textbf{48.23} \\
\toprule
\end{tabular}
\caption{Verification of the upper bound with \textit{GT\_Vertices} and the ablation study on pyramid feature, coordinate positional map, and number of vertices. All experiments are run on CAP dataset.}
\label{tab:upper} \vspace{-4mm}
\end{table}

\noindent\textbf{Ablation study.}
We conduct the ablation study to analyze the effectiveness of different algorithm modules. The  results are summarized in Table \ref{tab:upper}, where \textit{PyraFeat} for pyramid feature, \textit{CoordPos} for coordinate positional map, and \textit{N\_Vertices} for number of vertices/rays. (1) We investigate the impact of pyramid feature. In ablation, we only keep the $\frac{1}{4}$ scale of feature for vertex feature sampling, instead of four scales pyramid feature. From the second row of Table \ref{tab:upper} without the pyramid feature, the model performance drops significantly on all measures. This is because the boundary semantics in our work are affected by several factors including tumor size, tumor type and the slice position, and consequently the pyramid feature in multi-scale facilitates the integrated use of low-/mid-/high level cues naturally. (2) Adding the coordinate positional map demonstrates performance robustness improvement, as the sequential feature encodes the positional dependencies. (3) Due to the vital role of boundary vertices in our framework, we analyze the impact of different numbers of vertices (i.e., the number of rays). By comparing the third row versus the fifth row in Table \ref{tab:upper}, the setting of 30 vertices  yields  inferior  results than the default 90 vertices, implying that denser vertices are favored by our task.

\begin{table}[t] 
\centering
\footnotesize
\renewcommand\arraystretch{1.1}

\begin{tabular}{c|cccc}
\toprule
           & \textit{Sensitivity} &\textit{ Specificity} & \textit{AUC}   & \textit{J} \\ \hline \hline
UpperBound (Ground-truth) & 55.88       & 82.85       & 69.36 & 38.73      \\
Our prediction       & 67.64       & 72.85       & 70.25 & 40.50   \\  
\toprule
\end{tabular}
\caption{The evaluated results (\%) of logistic regression for predicting MVI from the capsular biomarker.} 
\label{tab:LR} 
\end{table}

{\bf Prognostic biomarker mining.} The inter-reader variability and the ambiguous annotation of capsular invasion in CT scans may undermine the performance since the reported results in Table. \ref{tab:main} are far less than perfect. This motivates us to further examine the effectiveness of our system on histopathology, which are widely considered as objective observation rather than subjective interpretation. The capsular biomarkers in the UpperBound setting uses features (pixel ratio of three classes, see Sec.~\ref{sec:prog}) calculated from the ground-truth CAP annotations while those biomarkers in ``our prediction'' uses features calculated from our model's prediction. Table. \ref{tab:LR} reports the correlation of the patient-specific capsular biomarkers and the MVI analyzed by logistic regression. It is surprising that our prediction slightly outperforms the upper bound in AUC, suggesting our predicted capsular biomarker performs comparable to radiologists' manual annotations for the clinically important task of prognostic MVI prediction.

\vspace{-2mm}

\section{Conclusion}
\vspace{-1mm}
Capsular invasion  on  tumor boundary has been clinically hypothesized of being correlated with the prognostic indicator MVI. In this paper, we present a novel quantitative computing framework on modeling the tumor boundary semantics of capsular invasion by disentangling this task to efficient spatial localization and sequential boundary semantics learning.  The detected tumor boundary semantics are directly converted into a prognostic biomarker that leads to a stronger statistical correlation with MVI than the version using human annotation. For the first time we interpret the boundary semantics as an effective tumor prognostic biomarker through objective computation, and provide an alternative non-invasive way to discover the subtle sign of prognostic vascular invasion.

{\small
\bibliographystyle{splncs04}
\bibliography{main}}

\end{document}